# Clustering validity based on the most similarity

R.Namayandeh,F.Didehvar,andZ.Shojaei

*Abstract*—One basic requirement of many studies is the necessity of classifying data. Clustering is a proposed method for summarizing networks. Clustering methods can be divided into two categories named model-based approaches and algorithmic approaches. Since the most of clustering methods depend on their input parameters, it is important to evaluate the result of a clustering algorithm with its' different input parameters, to choose the most appropriate one. There are several clustering validity techniques based on inner density and outer density of clusters that represent different metrics to choose the most appropriate clustering independent of the input parameters. According to dependency of previous methods on the input parameters, one challenge in facing with large systems, is to complete data incrementally that effects on the final choice of the most appropriate clustering. Those methods define the existence of high intensity in a cluster, and low intensity among different clusters as the measure of choosing the optimal clustering. This measure has a tremendous problem, not availing all data at the first stage. In this paper, we introduce an efficient measure in which maximum number of repetitions for various initial values occurs.

## I. INTRODUCTION

Clustering is a highly applicative method to categorizing many objects with several attributes into different classes in a way that the objects of the same class have similarity, and those that are broken down into different classes does not [١]. The existence of an edge between two vertices in a graph is defined as the criterion of similarity between them, in the other word, graph clustering is the task of grouping the vertices of the graph into clusters taking into consideration the edge structure of the graph in such a way that there should be many edges within each cluster and relatively few between the clusters [٢]. Clustering methods can be divided into two categories named as the model-based approaches and the algorithmic approaches. When the number of vertices and edges are large, it is more efficient to use model-based methods which have lower running time and memory. To modeling of networks, we have use random graphs. Random graphs are probable graphs in which the relationships between vertices are defined randomly. Erdos-Renyi is a random graph which generates this random edges by applying the Bernoulli distribution with parameter p. Although this model is known as a simple model, it is not used in real networks because of not containing primary features of those networks such as power-low degree. Therefore, alternative models such as mixture model were proposed to be used in the real networks [٣]. In these methods, the vertices' degrees of each cluster follow particular distributions. Then, some techniques such as maximum likelihood or expectation maximization are used for estimating the parameter of model. So, by considering the distribution and its' parameters of each cluster, we assign each vertex to its' appropriate cluster. In these methods, the sum of degrees of vertices for each cluster follows particular distributions. Then, some techniques such as Maximum likelihood or expectation maximization are used for estimating the parameter of the model, and by considering these parameters for each cluster, we can assign each vertex to its' appropriate cluster. Clustering is mostly an unsupervised process thus the evaluation of the clustering algorithms is very important. Since in the clustering process there are no predefined classes, it is difficult to find an appropriate metric for measuring whether the found cluster configuration is acceptable or not [١].

Several clustering validity approaches have been developed [٤,٥,٦,٧,٨]. The process of evaluating the results of a clustering algorithm is called cluster validity assessment.

These techniques for evaluating the result of the clustering algorithms are in accordance with inner density and outer density of clusters. In this paper, we introduce an efficient clustering such that maximum number of repetitions for various initial values occurred.

## II. CLUSTERING QUALIFYING BASED ON THE MOST SIMILARITY

One challenge of the large systems is to complete data gradually. In the other words, this type of systems use local clustering in which at the first level some data is received and the clustering occurred according only these data, and in each step, as the data completes gradually, the algorithm attaches each one of them to one of these clusters [٩]. So, since the initial data are highly determining in the way of clustering, defining the existence of high intensity in a cluster, and low intensity among different clusters, is not an appropriate measure to choose the optimal cluster, and the presence of all the data is not necessarily optimal clustering. In this paper, we introduce an efficient clustering in which the maximum number of repetitions for various initial values occurred. Therefore we have the following steps, in our algorithm:

**Step ١-** Finding the similarity degree between clusters and creating graph CC
**Step ٢-** clustering of the weighted graph $CC$:

## III. EXPERIMENTS AND RESULTS

In this section, in order to compare the proposed method with the similar methods, we use modularity measure (or Q measure) as the best available criteria [٨]. The comparison also has been applied on the email network [11] ,that shows the friendship among 1133 students of 10 faculties (Figure 1).

Q measure which is presented by Newman is defined as follows:

$$Q = \sum_{i=1}^{k}(e_{ii} - a_i^2)$$

In this equation, $k$ is the number of clusters, $e_{ij}$ is the half number of edges (with respect to undirected graph), $a_i$ be the fraction of all ends of edges that are attached to vertices in cluster $i$ and is calculated as:

$$a_i = \sum_j e_{ij}$$

The graph were clustered ten times by mixture model method [11] by different initial inputs. And we presented the result of our measure and other to compare them with optimum available solution. It should be noted that before any comparisons, the labels of clustering must be the same. So in the next section, we present a method for equalizing the label of clustering.

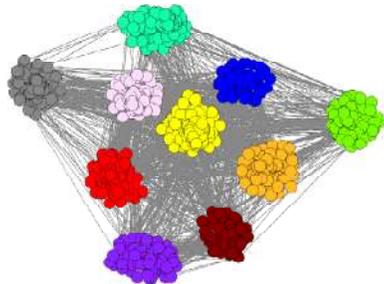

Figure 1. EMAIL graph with 1133 vertices and 3332 edge. the optimal clustering is available.

Where $C^*$ represents the optimum clustering, C represents current clustering and $O(x,y,z): N \times N \times N \rightarrow \{0,1\}$ is a function, where $x$ represents the label of input vertex, $y$ is the label of clustering and $z$ is the label of cluster and if the vertex $x$ is in the cluster $z$ of clustering $y$, then the value of $O(x,y,z)$ is equal 1 and otherwise it is equal 0.

After calculating the array $A$, we load $A$ as input parameter to the following algorithm .

*A. The result of experiments*

After equalizing the label of clusterings with the label of optimum clustering, now we can compute the error rate of clusterings. These values are given in TABLE 1.

TABLE 2 represents an adjacency matrix of a complete weighted graph that each vertex corresponds to a clustering. Then, the graph is clustered in to 8 clusters by WGC algorithm (TABLE 3). The best clustering according to equation (3) is No.2 clustering.

The results of the Q measure is given in TABLE 4 according to this measure; clustering No.8 is the best clustering which is also available in the clusters of the best clustering of proposed method, with the positive point that, according to TABLE 1, the best clustering selected by proposed method has less error rate than Q-measure.

In many networks such as web or social networks, obtaining the complete complementary information is impossible and as shown in Figure 3. The data collection is completed gradually. The clustering of the available data in the first stage is performed and at any stage of completing the data, entered data is assigned to an appropriate cluster. For this kind of clustering, the former measure of qualifying the clusterings is not suitable because if more results of clusterings have the same values and have less inner density and just one clustering has another value but have greater inner density than most values, then the result of qualifying the clustering selects that result with greater inner density and ignore the result of most clustering In other word, by using former measures for local clustering, we select the local optimum clustering.

However, to show this, the appropriate database did not found, so in this study, we generate a database that shown in Figure 2.

This data are the available data at first stage of Figure 3.

IV. CONCLUSION

In this paper, we proposed a new method to evaluate the result of clustering a system in which the data enters the set gradually. In this type of systems, it is common to use local clustering in which at the first level some data is received and the clustering occurred according only these data, and in each step, as the data completes gradually, the algorithm attaches each one of them to one of these clusters [9].
So, since the initial data are highly determining in the way of clustering, defining the existence of high intensity in a cluster, and low intensity among different clusters, is not an appropriate measure to choose the optimal cluster, and the presence of all the data is not necessarily optimal clustering.

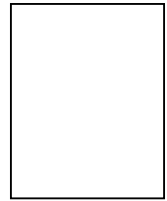

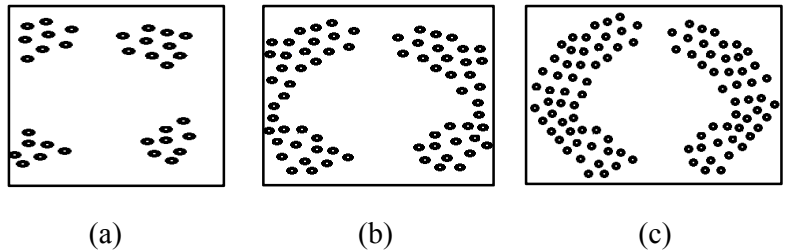

|  | (a) | (b) | (c) |
|---|---|---|---|

Figure ۲. Stages of data completing in ۳ phases (a),(b) and (c)

TABLE ۱.The error rate of each clustering compared to optimal clustering

| Clustering # | ۱ | ۲ | ۳ | ۴ | ۵ | ۶ | ۷ | ۸ | ۹ | ۱۰ |
|---|---|---|---|---|---|---|---|---|---|---|
| | ۱۲۷ | ۷۰ | ۹۰ | ۴۵۶ | ۴۷۰ | ۴۷۵ | ۷۳ | ۸۴ | ۴۲۰ | ۵۰۲ |

TABLE ۲.The similarity values of clusterings

|  | ۱ | ۲ | ۳ | ۴ | ۵ | ۶ | ۷ | ۸ | ۹ | ۱۰ |
|---|---|---|---|---|---|---|---|---|---|---|
| ۱ | 0 | 295103 | 256331 | 256133 | 271857 | 276110 | 290615 | 299755 | 264138 | 238910 |
| ۲ | 295103 | 0 | 306462 | 300546 | 330041 | 322390 | 351844 | 364421 | 316995 | 297961 |
| ۳ | 256331 | 306462 | 0 | 266670 | 283163 | 286792 | 305892 | 319234 | 281329 | 249949 |
| ۴ | 256133 | 300546 | 266670 | 0 | 288318 | 279170 | 300185 | 316771 | 278838 | 254273 |
| ۵ | 271857 | 330041 | 283163 | 288318 | 0 | 298401 | 317803 | 331593 | 298879 | 275383 |
| ۶ | 276110 | 322390 | 286792 | 279170 | 298401 | 0 | 332039 | 335560 | 298008 | 269108 |
| ۷ | 290615 | 351844 | 305892 | 300185 | 317803 | 332039 | 0 | 366598 | 318968 | 290417 |
| ۸ | 299755 | 364421 | 319234 | 316771 | 331593 | 335560 | 366598 | 0 | 327866 | 295330 |
| ۹ | 264138 | 316995 | 281329 | 278838 | 298879 | 298008 | 318968 | 327866 | 0 | 266450 |
| ۱۰ | 238910 | 297961 | 249949 | 254273 | 275383 | 269108 | 290417 | 295330 | 266450 | 0 |

TABLE ۳.The result of WGC algorithm on CC Graph

| Cluster# | ۱ | ۲ | ۳ | ۴ | ۵ | ۶ | ۷ | ۸ |
|---|---|---|---|---|---|---|---|---|
| | ۱ | ۲,۷,۸ | ۶ | ۵ | ۹ | ۳ | ۴ | ۱۰ |

TABLE ۴.The result of Q measure

| Clustering# | ۱ | ۲ | ۳ | ۴ | ۵ | ۶ | ۷ | ۸ | ۹ | ۱۰ |
|---|---|---|---|---|---|---|---|---|---|---|
| | 1735610.5 | 1176110.5 | 1726077 | 1544577 | 1318918.5 | 1236550 | 1066466 | 848295 | 1371728 | 1955792 |

TABLE ۵.The result of Dunn measure

| **Clustering** | ۱ | ۲ | ۳ | ۴ | ۵ | ۶ | ۷ | ۸ |
|---|---|---|---|---|---|---|---|---|
| **Value of *Dunn* measure** | ۰.۰۲۴۴ | ۰.۰۲۴۴ | ۰.۰۲۴۴ | ۰.۰۳۰۵ | ۰.۰۲۴۴ | ۰.۰۲۴۴ | ۰.۰۳۰۵ | ۰.۰۳۰۵ |

TABLE ۶. The result of WGC algorithm

|  | ۱ | ۲ | ۳ | ۴ | ۵ | ۶ |
|---|---|---|---|---|---|---|
| **Clustering By WGC algorithm** | ۱ | ۲ | ۳،۵،۶ | ۸ | ۴ | ۷ |